%% file: PaperForReview.tex
\documentclass[10pt,twocolumn,letterpaper]{article}

\usepackage[preprint]{cvpr}      

\usepackage{graphicx}
\usepackage{amsmath}
\usepackage{amssymb}
\usepackage{booktabs}
\usepackage{multirow}
\usepackage[T1]{fontenc}

\usepackage[pagebackref,breaklinks,colorlinks]{hyperref}

\usepackage[capitalize]{cleveref}
\crefname{section}{Sec.}{Secs.}
\Crefname{section}{Section}{Sections}
\Crefname{table}{Table}{Tables}
\crefname{table}{Tab.}{Tabs.}

\def\cvprPaperID{6580} 
\def\confName{CVPR}
\def\confYear{2022}

\begin{document}

\title{Efficient Classification of Very Large Images with Tiny Objects}

\author{Fanjie Kong, ~~~ Ricardo Henao\\
Department of Electrical and Computer Engineering \\
Duke University\\
{\tt\small \{fanjie.kong, ricardo.henao\}@duke.edu}
\and
{\tt\small }
}
\maketitle

\begin{abstract}
An increasing number of applications in computer vision, specially, in medical imaging and remote sensing, become challenging when the goal is to classify very large images with tiny informative objects. 
Specifically, these classification tasks face two key challenges: $i$) the size of the input image is usually in the order of mega- or giga-pixels, however, existing deep architectures do not easily operate on such big images due to memory constraints, consequently, we seek a memory-efficient method to process these images; and $ii$) only a very small fraction of the input images are informative of the label of interest, resulting in low region of interest (ROI) to image ratio.
However, most of the current convolutional neural networks (CNNs) are designed for image classification datasets that have relatively large ROIs and small image sizes (sub-megapixel).
Existing approaches have addressed these two challenges in isolation.
We present an end-to-end CNN model termed Zoom-In network that leverages hierarchical attention sampling for classification of large images with tiny objects using a single GPU.
We evaluate our method on four large-image histopathology, road-scene and satellite imaging datasets, and one gigapixel pathology dataset.
Experimental results show that our model achieves higher accuracy than existing methods while requiring less memory resources.
\end{abstract}

\section{Introduction}
Neural networks have achieved state-of-the-art performance in many image classification tasks \cite{krizhevsky2012imagenet}. 
However, there are still many scenarios where neural networks can still be improved.
Using modern deep neural networks on image inputs of very high resolution is a non-trivial problem due to the challenges of scaling model architectures \cite{tan2019efficientnet}.
Such images are common for instance in satellite or medical imaging.
Moreover, these images tend to become even bigger due to the rapid growth in computational and memory availability, as well as the advancements in camera sensor technology.
Specifically challenging are the so called \emph{tiny object image classification} tasks, where the goal is to classify images based on the information of very small objects or regions of interest (ROIs), in the presence of a much larger and rich ({\em non-trivial}) background that is uncorrelated or non-informative of the label.
Consequently, constituting an input image with a very low ROI-to-image ratio.
Recent work \cite{pawlowski2019needles} showed that with a dataset of limited size, convolutional neural networks (CNNs) have poor performance on very low ROI-to-image ratio problems.
In these settings, the input resolution is increased from typical image sizes, {\em e.g.}, $224 \times 224$ pixels, to {\em gigapixel} images of size ranging from $45,056 \times 35,840$ to $217,088 \times 111,104$ pixels \cite{CAMELYON}, which not only require significantly more computational processing power per image than a typical image given a fixed deep architecture, but in some cases, become prohibitive for current GPU-memory standards. 

\begin{figure}[t!]
\includegraphics[scale=0.27]{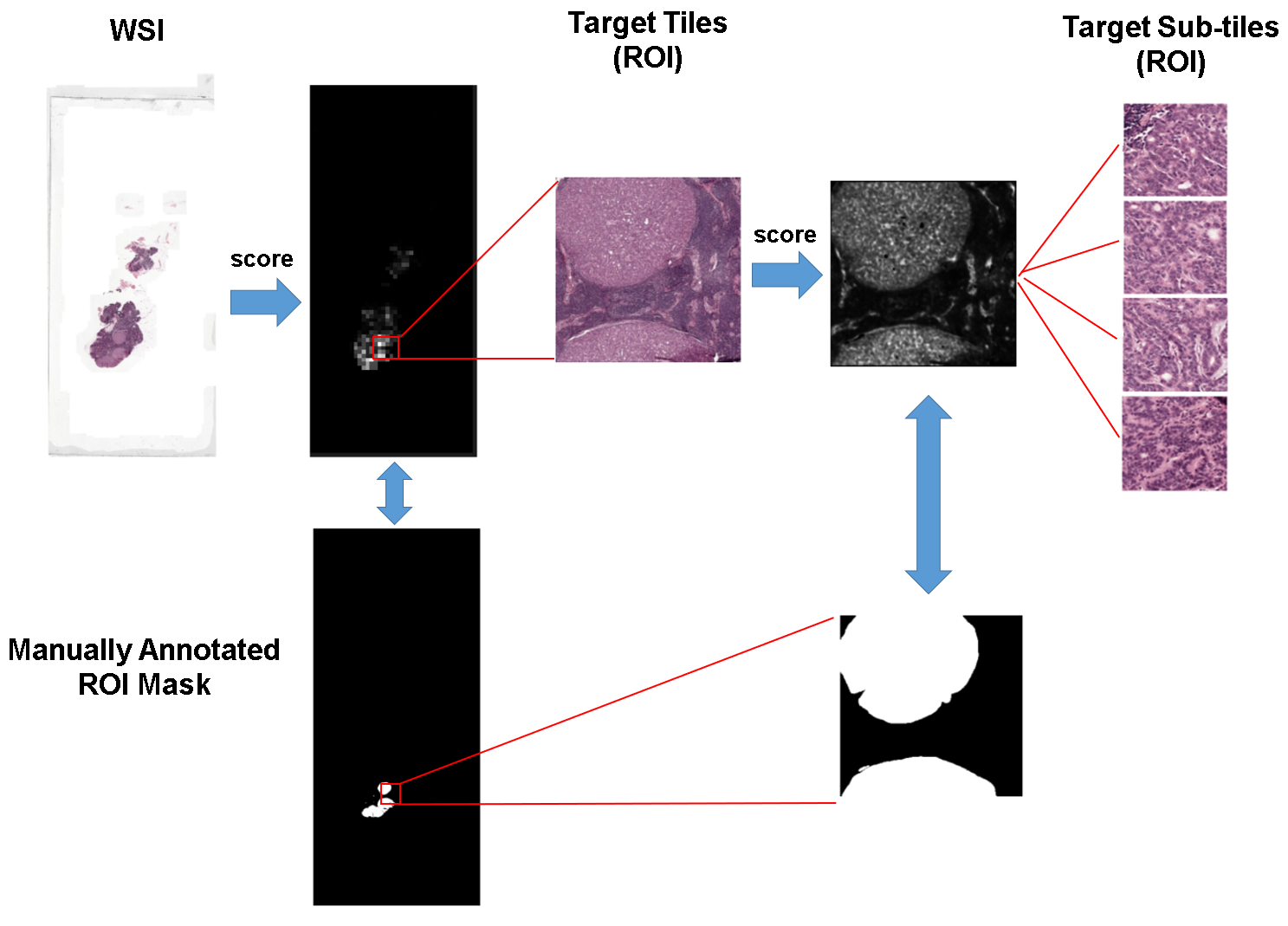}
\caption{Illustration of processing a typical WSI using our zoom-in strategy. We see that $i$) there are large regions with little information (mostly background), and $ii$) small informative regions have high-resolution details. Leveraging the above characteristics of WSI, we derive a method that gradually zooms-in to the ROI. The proposed approach first process the down-sampled WSI to sample the target tiles, and then repeats this procedure to sample target sub-tiles. The sampled sub-titles contain the fine-grained information for classification. The bottom images show that the manually annotated ROIs are captured by the proposed approach without the need for pixel level annotations.
}
\vspace{-4mm}
\label{fig:wsi}
\end{figure}

Figure~\ref{fig:wsi} shows an example of a gigapixel pathology image, from which we see that manually annotated ROIs (with cancer metastases), not usually available for model training, constitute a small proportion of the whole slide image (WSI).
Moreover, many tasks in satellite imagery \cite{christie2018functional} and medical image analysis \cite{CAMELYON} are still challenging due to the scarce methodology available for such big images.

Other recent works have addressed the computational resource bottlenecks associated with models for very large images by proposing approaches such as the streaming neural network \cite{Streaming_CNN} and gradient checkpoint \cite{Gradient_checkpoint}.
However, these methods do not take advantage of the characteristics of very large images in tiny object image classification tasks, {\em i.e.}, those in which only a small portion of the image input is informative for the classification label of interest.
Alternatively, other approaches use visual attention models to exploit these characteristics and show that discriminative information may be sparse and scattered across various image scales \cite{ATS,fu2017look,Hard_attention}, which suggests that in some scenarios, processing the entire input image is unnecessary, and specially true in tiny object image classification tasks.
For instance, \cite{ATS} leverages attention to build image classifiers using a small collection of tiles (image patches) {\em sampled} from the matrix of attention weights generated by an attention network.
Unfortunately, despite the ongoing efforts, existing approaches are either prohibitive or require severe resolution trade-offs that ultimately affect classification performance, for tasks involving very large (gigapixel) images.

The purpose of this work is to address these limitations simultaneously.
Specifically, we propose a neural network architecture termed {\em Zoom-In network}, which as we will show, yields outperforming memory efficiency and classification accuracy on various tiny object image classification datasets.
We build upon \cite{ATS} by proposing a two-stage {\em hierarchical} attention sampling approach that is effectively able to process gigapixel images, while also leveraging contrastive learning as a means to improve the quality of the attention mechanisms used for sampling.
This is achieved by building aggregated representations over a small fraction of high-resolution content (sub-tiles) that is selected from an attention mechanism, which itself leverages a lower resolution view of the original image.
In this way, the model can dramatically reduce the data acquisition and storage requirements in real-world deployments.
This is possible because low resolution views can be used to indicate which regions of the image should be acquired (attended) at higher resolution for classification purposes, without the need of acquiring the entire image at full resolution.
Moreover, we show that the proposed approach can be easily extended to incorporate pixel-level-annotations when available for additional performance gains.
Results on five challenging datasets demonstrate the capabilities of the Zoom-In network in terms of accuracy and memory efficiency.

\begin{figure*}[t!]
\centering
\includegraphics[scale=0.21]{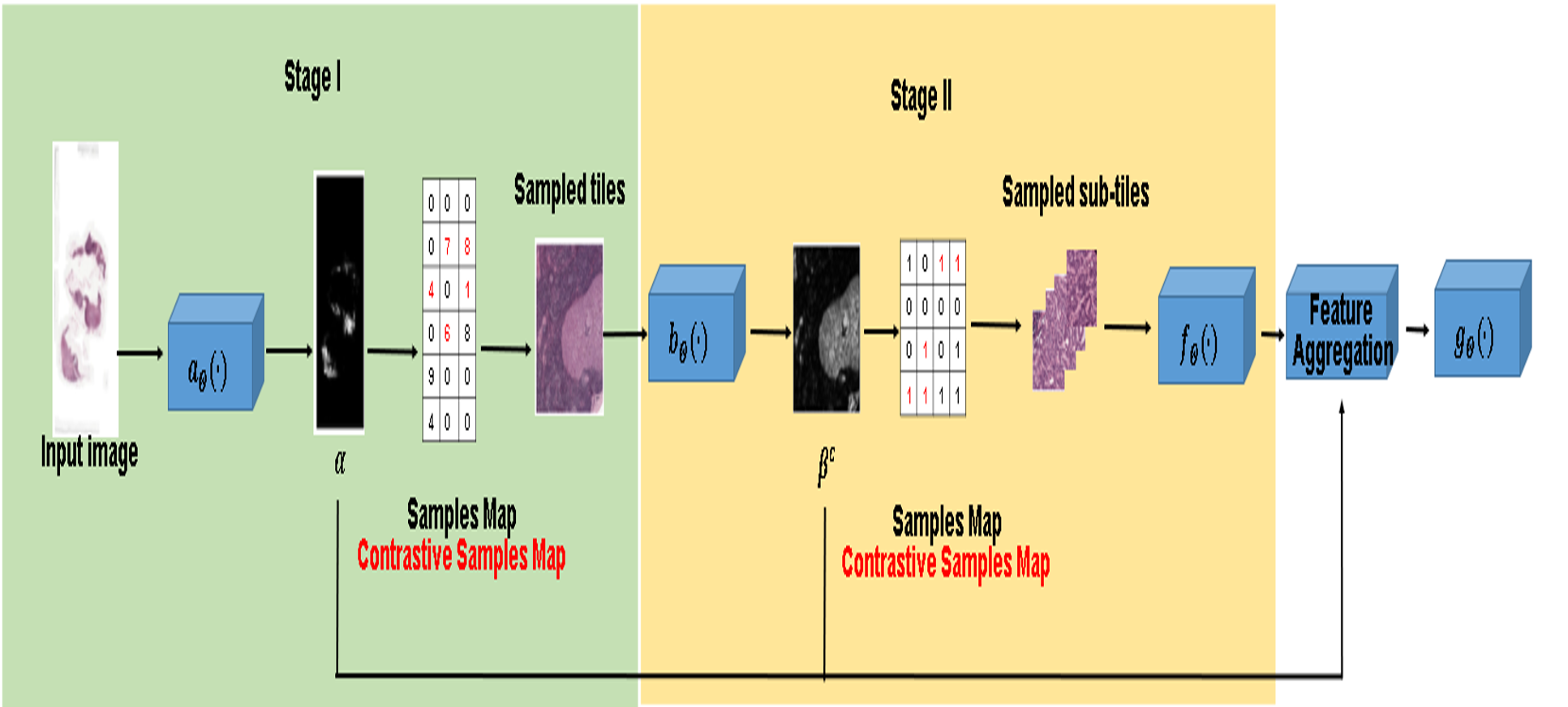}
\vspace{-3mm}
\caption{Illustration of the Zoom-In network.
In Stage I, attention network $a_\Theta(\cdot)$ generates an attention map for the input image down-scaled by $s_1$, from which $N$ tiles are sampled with replacement (see samples map).
In Stage II, attention network $b_\Theta(\cdot)$ generates an attention map for each selected tile and selects a sub-tile, thus $N$ sub-tiles are selected (without replacement).
Then all sub-tiles are fed to feature extractor $f_\Theta(\cdot)$, feature maps are aggregated using their corresponding attention weights, and predictions are obtained from aggregated features using a classification module $g_\Theta(\cdot)$.
Further, both attention maps are also used to draw contrastive samples with minimal computational overhead (during training).
}
\label{Model_summary}
\vspace{-3mm}
\end{figure*}

\section{Zoom-In Network}
\label{sec: methodology}
Below we present the construction of the proposed Zoom-In network model, which aims to efficiently process gigapixel images for classification of very large images with tiny objects.
We start by briefly describing the one-stage attention sampling method proposed in \cite{ATS}, which we leverage in our formulation.
Then, we introduce our strategy consisting in decomposing the attention-based sampling into two stages as illustrated in Figure~\ref{Model_summary}.
This {\em two-stage hierarchical sampling} approach enables computational efficiency without the need to sacrifice performance due to loss of resolution, when used in applications with very large images and small ROI-to-image ratios.
In the experiments, we will show that the Zoom-In network results in improved performance relative to existing approaches on several tiny object image classification datasets, and importantly, without the need for any pixel-level annotation.
\subsection{Two-stage Hierarchical Attention Sampling}\label{sc:stage2}
%
%
%
Let $T_{s_1}(x,c)$ denote a function that extracts a tile of size $h_1\times w_1$ from the input, full-resolution, image $x\in\mathbb{R}^{H\times W}$ corresponding to the location (coordinates) $c=\{i,j\}$ in a lower resolution view $V(x,s_1)\in\mathbb{R}^{h\times w}$ of $x$ at scale $s_1\in(0,1)$, so $h=\lfloor s_1H\rfloor$ and $w=\lfloor s_1W \rfloor$, where $\lfloor \cdot \rfloor$ is the floor operator.
More specifically, $T_{s_1}(x,c)$ maps $c$ to a location in $x$ via $\{\lfloor 1+(i-1)(W-1)/(w-1)\rfloor,\lfloor 1+(j-1)(H-1)/(h-1)\rfloor\}$, and returns a tile of size $h_1\times w_1$.
Note that
$i$) the map of locations between $V(x,s_1)$ and $x$ only depends on the size of $x$ ($H\times W$) and $s_1$ and not on the tile size ($h_1\times w_1$);
$ii$) $h_1,w_1>1/s_1$, to guarantee full coverage of $x$;
$iii$) this strategy requires to zero-pad $x$ on all sides by $\lfloor h_1/2 \rfloor$ and $\lfloor w_1/2 \rfloor$ pixels accordingly; and
$iv$) we have omitted the (color) channel dimension in $x$ and $V(x,s_1)$ for notational simplicity, however, we consider color images (with an additional dimension) in our experiments.
Then, let $\Psi_\Theta(x)=g_\Theta(f_\Theta(T_{s_1}(x,c)))$ be a neural network parameterized by $\Theta$ whose intermediate representation $z\in\mathbb{R}^K$ is obtained via feature extracting function $z=f_\Theta(T_{s_1}(x,c))$, \emph{e.g.}, a convolutional neural network (CNN).
Further, $g_\Theta(z)$ is a classification function also specified as a neural network and parameterized by $\Theta$.
We can provide $\Psi_\Theta(x)$ with an attention mechanism as follows
\begin{align}
    \alpha & = a_\Theta(V(x, s_1)):\mathbb{R}^{h\times w}\rightarrow \mathbb{R}^{h\times w } \label{eq:att} \\
    \Psi_\Theta(x) & = g_\Theta\left( \sum_{c\in C}\alpha_{c} f_\Theta(T_{s_1}(x,c)) \right) , \label{eq:psi}
\end{align}
where $\alpha$ is the matrix of attention weights such that $\sum_{c\in C} \alpha_c = 1$, $a_\Theta(V(x, s_1))$ is the attention function, also specified as a neural network, and $C$ (of length $|C|=h\cdot w$) is the collection of all index pairs for view $V(x,s_1)$.
In order to avoid computing the features $z$ from all $|C|$ tiles implied by view $V(x, s_1)$, which can be a very large number if $x$ is big, as in our pathology and remote sensing scenarios, \cite{ATS} proposed to leverage Monte Carlo estimation by only considering a small set of tiles from the original input image sampled via the attention function.
This strategy leverages that $\alpha$ defines a discrete distribution over the set of $|C|$ tiles.
Specifically, \cite{ATS} approximates \eqref{eq:psi} by sampling from \eqref{eq:att} via
\begin{align} \label{eq:mt1}
    \Psi_\Theta(x) \approx g_\Theta\left( \frac{1}{N}\sum_{c\in Q}f_\Theta(T_{s_1}(x,c)) \right) ,
\end{align}
where $Q$ is a collection of $N \ll |C|$ index pairs for view $V(x, s_1)$ drawn independently and identically distributed (iid) from the distribution defined by the attention weights, {\em i.e.}, $Q=\{(i, j)\sim \ a_\Theta(V(x, s_1)) | i=1, 2, ..., N\}$.
In \cite{ATS}, they consider tiles of size $h_1=w_1=27$, $s_1= 0.2$ and $N=10$ for the colon cancer dataset.
See Experiments below for additional details.

Using the approximation in \eqref{eq:mt1}, the attention mechanism uses a lower resolution view $V(x, s_1)$ of the original image $x$ for computing the attention distribution and outputs an aggregated feature vector by averaging over the features $\{z_n\}_{n=1}^N$ of a small amount of $N$ tiles.
Unfortunately, this approach is still prohibitive for gigapixel images because feasible combinations of $h_1$, $w_1$ and $s$ result in unrealistic memory needs for current GPU-memory standards.
Below, we introduce the proposed two-stage hierarchical sampling to improve the memory efficiency of attention sampling.

Multistage and hierarchical sampling strategies are often preferred in practice.
For instance, the cost of interviewing or testing people are enormously reduced if these people are geographically or organizationally grouped, thus sampling is performed within groups (clusters).
Such sampling design has many real-world applications such as household and mortality surveys, as well as high-resolution remote sensing applications \cite{household_survey,mortality_survey,xia2019high}.
Motivated by this idea, we design a two-stage hierarchical sampling approach to reduce memory requirements when processing very large, gigapixel, images without severe resolution trade-offs.

Specifically, let $V(x, s_2,c)\in \mathbb{R}^{u\times v}$ be a view of $T_{s_1}(x,c)$ at scale $s_2\in(0,1)$, so $u=\lfloor s_2h \rfloor=\lfloor s_1s_2H \rfloor$ and $v=\lfloor s_2w \rfloor=\lfloor s_1s_2W \rfloor$.
Further, we define a function $T_{s_2}(T_{s_1}(x,c),c')$ that extracts a sub-tile of size $h_2\times w_2$ at location $c'=\{i',j'\}$ in $V(x, s_2)$ from tile $T_{s_1}(x,c)$ at location $c=\{i,j\}$ in $V(x, s_1)$.
The mapping function $T_{s_2}(T_{s_1}(x,c),c')$ is defined similarly to $T_{s_1}(x,c)$, but returns tiles of size $h_2\times w_2$ instead of $h_1\times w_1$, and is such that $h_2<h_1$, $w_2<w_1$, and $h_2,s_2>1/s_2$.
Moreover, we can also define an attention mechanism for $V(x, s_2,c)$ as in \eqref{eq:att} as follows
\begin{align}
    \beta^c & = b_\Theta(V(x,s_2,c)):\mathbb{R}^{u\times v}\rightarrow \mathbb{R}_+^{u\times v} , \label{eq:att2}
\end{align}
where $\beta$ is the matrix of attention weights for the tile at location $c$ of $V(x,s_1)$ such that $\sum_{c'\in C'} \beta_c' = 1$, $b_\Theta(V(x,s_2,c))$ is the attention function, also specified as a neural network, and $C'$ (of length $|C'|=u\cdot v$) is the collection of all index pairs for view $V(x,s_2,c)$ of $T_{s_1}(x,c)$.
Provided that $\sum_{c\in C} \alpha_c = 1$ in \eqref{eq:att} and $\sum_{c'\in C'} \beta_c' = 1$ in \eqref{eq:att2}, it is easy to see that $\sum_{c\in C} \sum_{c'\in C'} a_\Theta(V(x,s_1))b_\Theta(V(x,s_2,c))=1$ and that the attention for location $c'=\{i',j'\}$ in $V(x, s_2,c)$ relative to the entire image $x$ is $\alpha_c\beta_{c'}$.
Consequently, we can rewrite \eqref{eq:psi} as
\begin{align} 
    \hspace{-2mm}\Psi_\Theta(x) & = g_\Theta\left( \sum_{c\in C}\alpha_{c} \hspace{-2mm}\sum_{c'\in C'} \beta_{c'}^c f_\Theta(T_{s_2}(T_{s_1}(x,c),c') ) \right) , \label{eq:psi2}
\end{align}
where now, the aggregated representation is a weighted average of all tiles of size $h_2\times w_2$ of $x$, and like in \eqref{eq:mt1}, we can approximate as
\begin{align} \label{eq:mt2}
    \Psi_\Theta(x) & \approx g_\Theta\left( \frac{1}{N}\sum_{c\in Q} f_\Theta(T_{s_2}(T_{s_1}(x,c),c')) \right) , 
\end{align}
where $c'\sim b_\Theta(V(x, s_s,c))$ is drawn iid from distribution $b_\Theta(V(x, s_s,c))$ for every location $c\in Q$.

Note that the approximation in \eqref{eq:mt2} uses {\em full-resolution} sub-tiles from $x$ that are drawn hierarchically from the two-level discrete distribution implied by $\alpha$ and $\{\beta^c\}_{c=1}^{|C|}$, which are obtained from {\em low-resolution} views $V(x,s_1)$ and $\{V(x,s_2,c)\}_{c=1}^{|C|}$.
Importantly, in practice we do not need to instantiate the tiles $T_{s_1}(x,c)$ but only $T_{s_2}(T_{s_1}(x,c),c')$, and the second-level attention matrix in \eqref{eq:att2} can be obtained as needed ({\em on the fly}).
However, this can cause computational inefficiency if multiple samples from the same location $c$ are selected for level-two sampling in \eqref{eq:mt2}.
Inefficiency occurs because such procedure will require to instantiate view $V(x,s_2,c)$ multiple times to obtain $\beta^c$ on a single model update (iterations), and then when a sub-tile is sampled multiple times when obtaining $f_\Theta(T_{s_2}(T_{s_1}(x,c),c')$.
We can mitigate the inefficiency by ordering the samples in $Q$ to prevent recalculating $\beta^c$, and we can reuse features $f_\Theta(T_{s_2}(T_{s_1}(x,c),c')$ for a given $c$ and $c'$ as needed.
Alternatively, we can avoid reusing sub-tiles by sampling locations $c'$ in \eqref{eq:mt2} {\em without replacement}.
However, such sampling strategy will not be iid and as a result, it will cause bias in the Monte Carlo approximation in \eqref{eq:mt2}.
Fortunately, using a formulation similar to that of \cite{ATS}, we can still obtain an unbiased estimator of the average (expectation) in \eqref{eq:mt2} from a non-iid sample, without replacement, by leveraging the Gumbel-Top-$k$ trick \cite{kool2019stochastic}, which is extended from the Gumbel-Max trick for weighted reservoir sampling \cite{efraimidis2006weighted}.
Specifically, from \eqref{eq:psi2} we can write
\begin{align} 
    & \mathbb{E}_{c'\sim b_{\Theta}(V(x, s_2, c))} [f_{\Theta}(T_{s_2}(T_{s_1}(x,c),c')) ] = \\
    & \hspace{38mm} \sum_{c'\in C'} \beta_{c'}^c f_\Theta(T_{s_2}(T_{s_1}(x,c),c') ) , \notag
\end{align}
from which we can see that the sum on the right is an unbiased estimator of the expectation on the left.
Alternatively, we can write
%
\begin{align}
    & \mathbb{E}_{c'\sim b_{\Theta}(V(x, s_2, c))} [f_{\Theta}(T_{s_2}(T_{s_1}(x,c),c')) ] = \label{eq:snorep} \\
    & \hspace{14mm} \sum_{c'\in C'} \sum_{i \neq c'} \beta_{c'}^c \frac{\beta_{i}^c}{1-\beta_{c'}^c}\left( \beta_{c'}^c f_\Theta(T_{s_2}(T_{s_1}(x,c),c')) \right) \notag \\
    & \hspace{28mm} + (1-\beta_{c'}^c)f_\Theta(T_{s_2}(T_{s_1}(x,c),i))) , \notag
\end{align}
where $\beta_{i}^c/(1-\beta_{c'}^c)$ is the attention weight for the $i$-th sub-tile reweighted to exclude sub-tile $c'$, which is equivalent to having already sampled it.
The proof of \eqref{eq:snorep} can be found in the Supplementary Material (SM).
We can then approximate \eqref{eq:psi2} like in \eqref{eq:mt2} but sampling without replacement using
%
%
\begin{align}\label{eq:mt2snorep}
    \begin{aligned}
    & \Psi_\Theta(x) \approx g_\Theta( \frac{1}{N}\sum_{i=1}^N \sum_{j=1}^{i-1} \beta_{c'_j}^{c_i} f_\Theta(T_{s_2}(T_{s_1}(x,c_i),c'_j))  \\ 
    & \hspace{14mm} + \left[ 1-\sum_{j=1}^{i-1} \beta_{c'_j}^{c_i} \right] f_\Theta(T_{s_2}(T_{s_1}(x,c_i),c'_j)) ) ,
    \end{aligned}
\end{align}
and $c'_j$ is sampled via
\begin{align}
    & c'_j \sim p(c'|c'_1,\ldots,c'_{j-1}) \propto \begin{cases} \beta_{i}^c & {\rm if} \ i \notin \{c'_1,\ldots,c'_{j-1}\} \\ 0 & {\rm otherwise} \end{cases} , \notag
\end{align}
where $p(c'|c'_1,\ldots,c'_{j-1})$ represents sampling location $c'_j$ without replacement, by having already sampled locations $c'_1,\ldots,c'_{j-1}$.

\paragraph{Memory requirements}
In practice, the memory requirements of the attention sampling model are determined by the model parameters, feature maps, gradient maps and workspace variables \cite{GPU_map}.
For neural-network-based image models, memory allocation is mainly dominated by the size of the input image, {\em i.e.}, $H$ and $W$. 
Specifically, the {\em peak memory} usage at inference for $N$ samples scales with $\mathcal{O}(s^{2}HW+Nh_2w_2)$ and $\mathcal{O}(s_{1}^{2}HW+N's_1^2s_2^2HW+Nh_2w_2)$ for both, the one-stage \cite{ATS} and the proposed two-stage hierarchical model. Here, we use $N'$ to denote the number of unique tiles in $Q$ and $s$ to indicate the scale of the view for the one-stage approach.
In fact, we can show that our model requires significantly less GPU memory than one-stage attention sampling by choosing $s_1 < s$ and $s_2 = s$.
Note that the number of selected tiles in the first stage decreases dramatically as the attention map is being optimized.
We use the term peak memory to refer to the worse case scenario.
Empirically, we have observed that the average number of selected tiles is $N'\approx N/2$.
A detailed analysis of memory requirements is presented in the SM.

\subsection{Efficient Contrastive Learning with Attention Sampling}\label{ICSIA}
Motivated by \cite{ki2020sample}, we introduce a contrastive learning objective for the proposed Zoom-In network consisting on encouraging the model to make predictions for cases ($y=1$), {\em e.g.}, images with cancer metastases (see Experiments for details), but using sub-tiles with low attention weights while {\em inverting} the image labels ($y=1\to 0$).
Conveniently, we can generate these (negative) contrastive samples without the need for additional modules or model parameters. 

Specifically, we leverage the existing attention functions in \eqref{eq:att} and \eqref{eq:att2}.
To generate the contrastive feature vectors for image $x$ such that $y=1$, we first sample (with replacement) tile locations via $1-a_\Theta (V(x, s_1))$ similar to \eqref{eq:att}.
Then, we sample $N$ sub-tiles via $1-b_\Theta(V(x, s_2,c))$ without replacement similar to \eqref{eq:att2}.

The sampled contrastive sub-tiles are passed through the feature network and then processed by the classifier to make predictions $\Psi_\Theta(x|y=1)$ using \eqref{eq:mt2snorep}, where the conditioning $y=1$ is used to emphasize that we use images $x$ of class $y=1$ as contrastive examples.
In general, the number of contrastive examples (per training batch) is equal to the number of samples such that $y=1$.
For these contrastive sample, we optimize the following objective, $\mathcal{L}_{\rm con}(\Psi_\Theta(x|y=1)) = \sum_n - \log(1-\Psi_\Theta(x_n|y_n=1))$.
Note that $\mathcal{L}_{\rm con}(\Psi_\Theta(x|y=1))$ encourages contrastive samples for images $x$ with label $y=1$ to be predicted as $y=0$.
In multi-class scenarios, this contrastive learning approach can be readily extended by letting one of the classes be the reference, or in general, by using a complete, cross-entropy-based contrastive loss, in which contrastive samples are generated for both classes, {\em i.e.}, $y=\{0,1\}$, instead of just one class (half the cross-entropy loss) as in our case.

\section{Related Work}
\label{sec: related work}
Below we discuss existing research work on classification of very large images with tiny objects, the most relevant body of work being on attention-based models, and general efforts toward computational efficiency for image classification models.

\noindent{\bf Tiny object classification}
Recent works have studied CNNs under different noise scenarios, either by performing experiments where label noise is introduced artificially \cite{zhang2016understanding,arpit2017closer}, or by directly working with noisy labels and annotations \cite{mahajan2018exploring,han2018co}.
While it has been shown that large amounts of label noise hinders the generalization ability of CNNs \cite{zhang2016understanding,arpit2017closer}, it has been further demonstrated that CNNs can mitigate this label-corrupting noise by increasing the size of the data used for training~\cite{mahajan2018exploring}, tuning the parameters of the optimization procedure ~\cite{jastrzkebski2017three}, or re-weighting input training samples \cite{han2018co}.
However, all of these works focus on label corruption but do not consider the case of noiseless labels or label assignments with low-noise, in which alternatively, the region of interest (ROI) associated with the label is small or tiny, relative to the size of the image.
Purposely, \cite{pawlowski2019needles} analyzed the capacity of CNNs in precisely this context, {\em i.e.}, that of tiny object image classification tasks.
Their results indicate that by using a training dataset limited in size, CNNs are unable to generalize well as the ROI-to-image ratio of the input decreases.
Typically, the object associated with the label occupies a dominant portion of the image.
However, in some real-world applications, such as medical imaging, remote sensing or traffic signs recognition, only a very tiny fraction of the image informs their labels, leading to a low ROI-to-image ratios.

\noindent{\bf Attention}
This technique has a long history in the neural networks literature \cite{itti1998model}.
In the modern era of deep learning, it has been used very successfully in various problems \cite{denil2012learning, fu2017look, zheng2019looking, wang2017zoom}.
Two main classes of attention mechanisms include: {\em soft attention}, which estimates a (continuous) weight for every location of the entire input \cite{MIL}, and {\em hard attention} which selects a fraction of the data, {\em e.g.}, a ROI in an image, for processing  \cite{Hard_attention}, which is a harder problem that resembles object detection, but without ground-truth object boundaries.
Note that the attention in \cite{ATS, MIL,brendel2019approximating} is defined as the weights of a bag of features from an image.
Our formulation can also be interpreted in the same way, since $\alpha$ is the attention on bag of features of tiles and $\beta^{c}$ is the attention on bag of features of sub-tiles.

\noindent{\bf Computational efficiency}
There are multiple ways to control the computational cost of deep neural networks.
We categorize them into four groups:
$i$) compression methods that aim to remove redundancy from already trained models \cite{yu2018nisp};
$ii$) lightweight design strategies used to replace network components with computationally lighter counterparts \cite{jaderberg2014speeding};
$iii$) partial computation methods selectively utilize units of a network, thus creating forward-propagation paths with different computational costs \cite{larsson2017ultra};
and $iv$) reinforcement learning and attention mechanisms that can be used to selectively process subsets of the input, based on their importance for the task of interest \cite{ramapuram2018variational, levi2018efficient,uzkent2020learning, ATS, cordonnier2021differentiable}.
The latter being the strategy we consider in the proposed Zoom-In architecture.

\noindent{\bf In-sample contrastive learning}
Conventional deep neural networks lack robustness to out-of-distribution data or naturally-occurring corruption such as image noise, blur, compression and label corruption.
Contrastive learning \cite{ki2020sample} has demonstrated great success with learning in noisy scenarios, {\em e.g.}, corrupted ImageNet \cite{khosla2020supervised}.
Here, we aim to leverage contrastive learning to mitigate the performance loss caused by images with low ROI-to-image ratio.
Hence, the built-in attention mechanism facilitates the in-sample contrastive learning because contrastive samples can be obtained using the same attention mechanisms without the need for additional model components or parameters.

\noindent{\bf Weakly supervised training on gigapixel images}
Recent work has demonstrated that deep learning algorithms have the ability to predict patient-level attributes, {\em e.g.}, cancer staging from whole slide images (WSIs) in digital pathology applications \cite{lee2018robust}.
Because these images are so large and no prior knowledge of which subsets of the image (tiles) are associated with the label, such task is known as weakly supervised learning \cite{campanella2019clinical,lu2021data}.
Specifically, the model has to estimate which regions within the image are relevant to the label, so predictions can be made using information from these regions alone; not the whole image.
Importantly, with current hardware architectures, WSIs are too large to fit in GPU memory, so one commonly used technique is to build a model to select a subset of patches from the image \cite{zhu2016deep,naik2020deep}.
An alternative approach consists in using the entire WSI but in a compressed, much smaller, representation, at the cost of losing fine-grained details that may be important \cite{tellez2019neural}.
Building representations that aggregate features from selected image regions (tiles or patches) are also alternative approaches \cite{campanella2019clinical,courtiol2018classification}.
We consider the performance of these approaches relative to the proposed Zoom-In network in the experiments.
%

\section{Experiments}
\label{sec: experiments}
We evaluate the proposed approach in terms of accuracy and GPU memory requirements. In the results below, Zoom-In Network refers to the proposed method with a lightweight LeNet backbone, and Zoom-In Network (Res) refers to our method using a ResNet16 backbone. The details of the model architecture are presented in the SM.
Moreover, we highlight the ability of the model to attend to a small amount of full-resolution sub-tiles (ROIs) of the image inputs, which results in a significantly reduced peak GPU memory usage and superior test accuracy relative to the competing approaches.
We consider methods that can handle large images as inputs, {\em e.g.}, attention sampling models (ATS) \cite{ATS}, Differentiable Patch Selection (Top-K) \cite{cordonnier2021differentiable}, BagNet \cite{pawlowski2019needles}, EfficientNet \cite{tan2019efficientnet} and streaming CNNs \cite{Streaming_CNN}.
Simultaneously, we also compare our model with methods that apply strategies similar to zoom-in {\em e.g.}, PatchDrop \cite{uzkent2020learning} and RA-CNN \cite{fu2017look}.
For the peak memory usage, we report inference memory in Mb per sample, {\em i.e.}, for a batch of size 1. We also report the floating point operations (FLOPs) and run time when inferring a single image.
Details of the model architecture and some hyper-parameters not specified in each experiment, as well as an ablation study examining the impact on performance of $N$, $\lambda$, and using contrastive learning are presented in the SM.
In-sample contrastive learning is applied after training without it for 10 epochs, and the entropy regularization parameter (for our model and ATS) is set to $\lambda = 1e^{-5}$.

\begin{table}[t!]
\caption{Test set results for colon cancer, NeedleCamelyon and fMoW data. Memory denotes average peak memory per sample usage at inference.
} 
\vspace{-3mm}
\centering 
\small
\resizebox{1.0\columnwidth}{!}{
\begin{tabular}{c|c c c r} 
    \multicolumn{5}{c}{Colon Cancer}  \\
    \hline
    & Accuracy & FLOPs  & Memory  &  Time \\ 
    Method   & (\%) &  (B) &  (Mb) &   (ms)       \\
    \hline
    PatchDrop & 81.0 & 75.29 & 520.44 & 12.33 \\
    RA-CNN & 86.4 & 135.88 & 4432.46 & 38.74 \\
	CNN & 90.8 $\pm$ 1.2 &  1.83 & 235.68 & 7.62  \\  
	ATS & 90.7  $\pm$ 1.4 & 0.24  & 15.83 & \bf 2.81 \\ 
    Zoom-In Network (ours) & \bf { 95.0 $\pm$ 2.6} & \bf 0.24 & \bf { 2.55} & 3.20 \\
    \hline
    \multicolumn{5}{c}{NeedleCamelyon} \\
    \hline
    & Accuracy & FLOPs  & Memory  &  Time \\ 
    Method   & (\%) &  (B) &  (Mb) &   (ms)       \\
    \hline
	BagNet   & 70.0 &  222.72 & 3914.81 & 12.90 \\  
	ATS  & 72.5 & 1.66 & 37.97 & \bf 9.27 \\ 
	Zoom-In Network (ours) & 76.0 & \bf 0.52 & \textbf{11.78} & 10.20 \\
	Zoom-In Network (Res) (ours)  & \bf 78.1 &  0.84 &  14.22 & 11.42 \\ 
	\hline
    \multicolumn{5}{c}{Traffic Signs Recognition} \\
    \hline
    & Accuracy & FLOPs  & Memory  &  Time \\ 
    Method   & (\%) &  (B) &  (Mb) &   (ms)       \\
    \hline
    EfficientNet-B0 s0.5 &  65.9 & 4.82 & 673.39 & 27.06\\
    EfficientNet-B0 &  79.1 & 19.26 & 2229.59 & 34.88 \\ 
	ATS-10 &  90.5 & 1.43 & 54.51 & 10.3 \\
	Top-K-10 & 91.7 & 1.43 & 53.28 & \bf 9.8 \\
	Zoom-In Network (ours) &  {91.2} & \bf 0.79 & \bf 12.65 &  12.28 \\
	Zoom-In Network (Res) (ours) & \bf {92.6} &  1.18 &  15.83 & 13.16 \\ 
    \hline
	\multicolumn{5}{c}{Functional Map of the World} \\
    \hline
    & Accuracy & FLOPs  & Memory  &  Time \\ 
    Method   & (\%) &  (B) &  (Mb) &   (ms)       \\
    \hline
	EfficientNet-B0 & 70.2 & 8.22 & 1404.09 & 22.72 \\
	Zoom-In Network (ours)  &  {72.9} & \bf 1.85 & \bf {10.81} & \bf 11.47 \\ 
	Zoom-In Network (Res) (ours)  & \bf {74.3} &  2.24 &  13.53 &  12.25 \\ 
\end{tabular}}
\label{table:histo} 
\vspace{-3mm}
\end{table}

{\bf Datasets}
We focus on datasets that have relatively large image sizes and feature tiny ROI objects that are scattered in large backgrounds, unlike natural images, in which objects are (usually) in the middle of the image due to the fact that photographers tend to center images around the object of interest (target) \cite{touvron2019fixing}. Consequently, in the experiments, we do not consider datasets such as ImageNet, iNaturalist and COCO because in these the ROI-to-image ratio is close to 1 and also because image sizes are relatively small relative to some of the other datasets considered and described below.

We present experiments on five datasets:
$i$) the colon cancer dataset introduced in \cite{colon_cancer} aims to detect whether epithelial cells exist in a Hematoxylin and Eosin (H\&E) stained images.
This dataset contains 100 images of dimensions 500 × 500.
The images originate both from malignant and normal tissue, and contain approximately 22,000 annotated cells.
Following \cite{MIL,ATS}, we treat the problem as a binary classification task where the positive images are those containing at least one cell belonging to the epithelial cell class.
$ii$) The NeedleCamelyon dataset \cite{pawlowski2019needles} is built from cropped images from the original Camelyon16 dataset with specified ROI-to-image ratios.
Specifically, we generate datasets for ROI-to-image ratios in the range of $[0.1,1]\%$, and we crop each image with size of $1,024 \times 1,024$ pixels.
Positive examples are created by taking 50 random crops from every annotated cancer metastasis area if the ROI-to-image ratio falls within the range $[0.1,1]\%$.
Negative examples are taken by randomly cropping normal whole-slide images and filtering out image crops that mostly contain background.
Further, we ensure the class balance by sampling an equal amount of positive and negative crops.
$iii$) Traffic Signs Recognition dataset \cite{larsson2011using} consists of over $20, 000$  road scene images with a size of $960 \times 1280$. Here, we use the same subset as \cite{ATS, cordonnier2021differentiable}. The task is to classify whether the road-scene image contains a speed limit sign ($50$, $70$ or $80$ km/h) or not. The subset used in \cite{ATS,cordonnier2021differentiable} and our experiments includes $747$ images for training and $684$ images for testing.
$iv$) The Functional Map of the World (fMoW) dataset \cite{christie2018functional} aims to classify the functional purpose of buildings and infrastructures and land-use from high-resolution satellite images.
The approximate range of image size in this dataset is $500 \times 500$ to $9,000 \times 9,000$ pixels.
In our experiment, we extract a class-balanced subset from the original fMoW dataset to further illustrate the proposed method is applicable beyond digital pathology images. Our constructed subset consists of $15,000$ training images and $9,571$ test image from 10 classes. More details of constructing the subset is provided in SM.
$v$) Finally, we utilize the Camelyon16 dataset to further demonstrate the utility of our model on gigapixel images.
This dataset contains 400 WSIs, 270 WSIs with pixel-level annotations, and 130 unlabeled WSIs as test set.
We split the 270 slides into train and validation sets; for hyperparameter tuning.
Typically, only a small portion of a slide contains biological tissue of interest, with background and fat encompassing the remaining areas, {\em e.g.}, see Figure~\ref{fig:wsi} for a typical WSI (with pixel-level annotations).




{\bf Colon cancer}
%
%
%
We use the same experimental setup as \cite{ATS,MIL}, namely, 10-fold-cross-validation, and five repetitions per experiment.
%
The approach most closely related to ours is the one-stage attention sampling model in \cite{ATS}.
We refer to this method as ATS-$N$, where $N=10$ indicates the number of tiles drawn from the attention weights, and we set $s=s_2$ in all experiments.
The value of $N$ was selected to maximize performance.
To showcase the advantages of our model compared to traditional CNN approaches, we also include a ResNet \cite{he2016deep} with 8 convolutional layers and 32 channels as naive baseline.
%
For our model, we set $s_1 = 0.1$, $s_2 = 0.2$, $N=10$, and $h_2 = w_2 = 27$.

Results are summarized in Table~\ref{table:histo}. 
The proposed two-stage attention sampling model results in approximately 4.3\% higher test accuracy than (one-stage) ATS-10; presumably due to its ability to better focus on informative regions (sub-tiles) of the image via the hierarchical attention mechanism.
Moreover, the baseline (CNN) and ATS-10 require at least $\times 90$ and $\times 6$ more memory relative to the Zoom-In network, respectively.
Alternatively, PatchDrop \cite{uzkent2020learning} and RA-CNN \cite{fu2017look} not only underperform the base CNN but also have higher memory requirements.
The memory efficiency of the proposed approach is justified by the way in which the image is processed as a small collection of sub-tiles, thus resulting in a substantially reduced forward pass cost relative to the CNN and ATS-10 models. The FLOPs of ATS and Zoom-In Network are comparable because in the colon cancer experiments, the feature extractor $f_\Theta(\cdot)$ dominates the FLOP count. Since ATS and Zoom-In Network feed the same number of patches into  $f_\Theta(\cdot)$, their resulting FLOPs in this step is the same. In fact, this step contributes $0.235$ B FLOPs (96\%).
Further, our model takes slightly more run time due to the implementation inefficiency when extracting tiles and sub-tiles.

{\bf NeedleCamelyon}
%
Note that the NeedleCamelyon dataset uses larger images ($1,024 \times 1,024$) compared to the ones (up to $512 \times 512$) used in \cite{pawlowski2019needles} to better showcase the abilities of the models considered.
%
%
Following \cite{pawlowski2019needles}, we split the NeedleCamelyon dataset into training set, validation set, test set with a ratio of $60$:$20$:$20$.
The number of images for each set are $6,000$, $2,000$ and $2,000$, respectively.
Positive and negative samples are balanced in each set.

We compare our model with ATS-30 and an existing CNN architecture used for a similar NeedleCamelyon dataset in \cite{pawlowski2019needles}.
BagNet \cite{brendel2019approximating} is a CNN model extracts features at tile-level, which is efficiently used in NeedleCamelyon experiments in \cite{pawlowski2019needles}.
We use the BagNet as the CNN baseline in our experiment.
For our model, we set $s_1 = 0.25$, $s_2 = 0.5$, $N=30$, and $h_2=w_2 = 32$.
Due to the much smaller ROI-to-image ratio of NeedleCamelyon relative to the colon cancer dataset, we set a larger $s_1$ and $s_2$ to ensure the down-sampling will not wash out the discriminative information.
We also increase $N$ since the number of attention weights with large values is larger in this case.

Table \ref{table:histo} shows performance for the proposed model and the baselines.
We observe that for larger images, the Zoom-In network has better GPU memory use at inference time because much less sub-tiles at scale $s_2$ tend to be instantiated.
In terms of test accuracy, the proposed model results in approximately $3.5\%$ higher test accuracy than one stage ATS-30 and $6.0\%$ higher than the BagNet baseline in this experiment.
Moreover, BagNet and one-stage attention sampling require at least $\times 500$ and $\times 7$ more memory, respectively, compared to the proposed approach. Here, we can see our Zoom-In Network consumes drastically less FLOPs than ATS. This is because the Zoom-In Network requires less pixels of the original input images to select a comparable amount of high resolution ROI patches for prediction relative to ATS.

\begin{figure}[t!]
\centering
\includegraphics[scale=0.38]{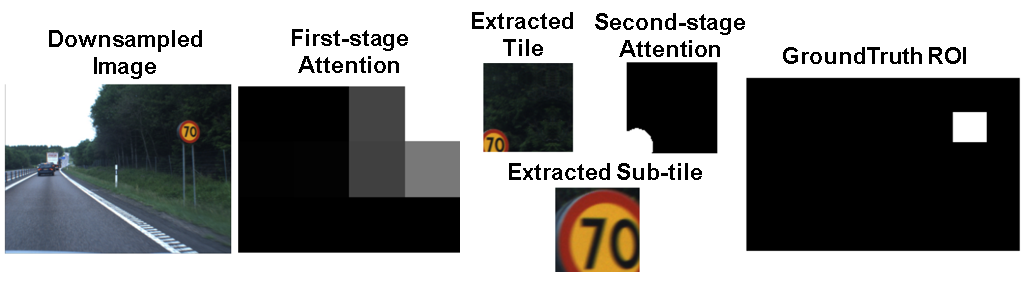}

\vspace{-2mm}
\caption{Illustration of the intermediate results for the Traffic Sign dataset using the Zoom-In Network. We show the tile and sub-tile with the highest first-stage and second-stage attention. More intermediate results are included in the SM.
}
\vspace{-5mm}
\label{fig:amap}
\end{figure}

{\bf Traffic Sign Recognition}
We compare the Zoom-In Network with a traditional CNN (EfficientNet-B0 \cite{tan2019efficientnet}) and the recently published one-stage zoom-in methods (ATS \cite{ATS}, TopK \cite{cordonnier2021differentiable}). For EfficientNet-B0, we use both the original resolution images and images downsampled by half (denoted as s0.5) as inputs, to show the limitations of traditional CNNs. For ATS and TopK, we use the same hyperparameter settings from \cite{ATS, cordonnier2021differentiable}. For our Zoom-In Network, we tried two types of the backbones(LeNet and ResNet16) and we set $s_1 = 0.125$, $s_2 = 0.3$, $N=10$, and $h_2=w_2 = 100$. The details of the network architecture are shown in the SM. 

In Table \ref{table:histo}, we can see that our Zoom-In model achieves the highest accuracy and lowest GPU memory consumption among all tested methods. Also, our model requires less FLOPs by using less number of input pixels than ATS and Top-K.
The merits of our model originate from accurate target object localization achieved by the attention mechanism and contrastive learning, as well as efficient GPU memory usage. Examples of attention maps for Traffic Sign dataset similar to those in Figure~\ref{fig:amap} are shown in the SM.

\begin{table}
\caption{Results on Camelyon16 data. Memory denotes average peak memory per sample at inference.} 
\vspace{-3mm}
\centering 
\resizebox{1.0\columnwidth}{!}{
\begin{tabular}{c c c r} 
  & Pixel-level & Accuracy & Memory \\
Method & Annotation &  (\%) &  (Mb) \\
\hline 
	Streaming CNN \cite{Streaming_CNN} & No & 70.6 &  3,256.29 \\ 
	CLAM \cite{lu2021data} & No & 78.0 & 206.88 \\
	MIL \cite{campanella2019clinical} & No & 79.9 & 140.68  \\
	MRMIL \cite{li2021multi} & No & 81.1 & 568.00  \\
	Zoom-In Network & No &\bf {81.3} & \bf {71.76} \\
	Zoom-In Network (Res) & No &\bf {82.6} & \bf {71.76} \\
	\hline
	Zoom-In Network & Yes & 88.2 & \bf {71.76}\\ 
	Zoom-In Network(Res) & Yes & 90.8 & \bf {71.76}\\ 
 Winning model \cite{bejnordi2017diagnostic}
   & Yes & 92.2 & 395.77 \\  
\end{tabular}}
\label{table:CAM16_result} 
\vspace{-3mm}
\end{table}

{\bf Functional Map of the World}
Functional Map of the World (fMoW) consists of a large amount of high-resolution RGB images of various sizes ranging from $500 \times 500$ to $9,000 \times 9,000$.
Following \cite{Hard_attention}, we choose EfficientNet-B0 \cite{tan2019efficientnet} as the baseline model.
EfficientNet-B0 effectively scales with large images and has been proved to serve as a good baseline model on the fMoW dataset in \cite{Hard_attention}.
For EfficientNet-B0, the input images are resized to $896 \times 896$ and all other hyper-parameters are consistent with \cite{Hard_attention}.
For our model, we set $s_1 = 0.25$, $s_2 = 0.5$, $N=30$, and $h_2=w_2 = 50$.

Results in Table \ref{table:histo} show that Zoom-In Network surpasses EfficientNet-B0 in accuracy and memory consumption at inference  time. Examples of attention maps for fMoW similar to those in Figure~\ref{fig:amap} are shown in the SM.

{\bf Camelyon16}
This is a gigapixel dataset consisting of whole-slide images (WSIs) with sizes ranging from $45,056\times 35,840$ to $217,088\times 111,104$.
%
The objective here is to predict whether a WSI contains cancer metastases. 
%
As we described in the Related Work Section, existing works have attempted to train CNNs on very large images with only image-level labels, {\em i.e.}, via weakly supervised training on gigapixel images.
We consider the streaming CNN \cite{Streaming_CNN}, CLAM \cite{lu2021data}, MIL \cite{campanella2019clinical} and MRMIL \cite{li2021multi}, which we previously briefly described.
Details of these baselines are presented in the SM. 
Further, we also evaluate the model for the scenario in which pixel-level annotations (ROIs) are available.
Here, we compare our results to the winning model of the Camelyon16 challenge \cite{bejnordi2017diagnostic}. 
%
For our model, we set $s_1 = 0.03125$, $s_2 = 0.125$, $N=100$, and $h_2 = w_2 = 50$.

In Table \ref{table:CAM16_result}, we see that our Zoom-In network achieves the highest test accuracy when pixel-level annotations are not available.
Even when there are pixel-level annotations, our model yields a test accuracy close to the winning model of the Camelyon16 challenge, without the need for comprehensive tuning and handcrafted features ({\em i.e.}, the minimum surrounding convex region, the average prediction values, and the longest axis of the lesion area).
For the memory comparison, all methods require substantially more memory than the proposed approach, notable $\times 8$ more than MRMIL which has comparable performance.
Moreover, we also consider the situation in which the pixel-level annotations are available. The extension to leverage pixel-level annotations is described in the SM. The performance of the proposed model is relatively close to the winning model \cite{bejnordi2017diagnostic}, which indicates that the proposed approach is flexible and also accurate when we have the manually annotated ROIs.
Additional details including attention maps similar to that in Figure~\ref{fig:amap} are provided in the SM. 

We also analyzed the correlation of the attention weights generated by the Zoom-In network with the ground-truth metastases-to-tile ratio obtained from pixel annotations.
Specifically, the proportion of each sub-tile of size $h_2\times w_2$ that is covered by cancer metastases.
The Spearman correlation coefficient for all the tiles with pixel-level annotations is $\rho=0.3570$, indicating a good agreement.
Note that these attention weights are obtained from the model trained without pixel-level annotation information.
In the SM, we visually present these correlations as scatter plots (attention weights {\em vs.} metastases-to-tile ratios).

From all the results above, we see our Zoom-In Network reduces the GPU memory usage significantly. The reasons why we pursue a low GPU memory consumption are: $i)$ GPUs with high memory are expensive and not widely available for applications in practice. A model using less GPU at inference time can be deployed with less expense; $ii)$ the highly memory-efficient model allows training and inferring images larger than gigapixels possible; $iii)$ With the growing use of neural network on mobile/edge devices, it is very important to develop light GPU memory consumption models to allow locally running deep learning service on mobile/edge devices.
\section{Discussion}
%
We presented the Zoom-In network that can efficiently classify very large images with tiny objects.
We improved over the existing CNN-based baselines both in terms of accuracy and peak GPU memory use, by leveraging a two-stage hierarchical sampling strategy and a contrastive learning objective.
We also considered the scenario in which pixel-level annotations (segmentation maps) are available during training.
In the experiments, we demonstrated the advantage of the proposed model on five challenging classification tasks.
We note that the images in the Camelyon16 dataset are all gigapixel in size and that we are likely the first ones training an end-to-end deep learning model for them.
Our model achieved the best accuracy when pixel-level annotations are not available, while also using a small amount of GPU memory, which allows for training and inference on full-resolution gigapixel images using a single GPU.
One limitation of the proposed model is the need to specify the number of sub-tile samples $N$, which can be potentially estimated from data. Moreover, exploring the impact of additional steps in the zoom-in hierarchy is also of interest but left as future work.


{\small
\bibliographystyle{ieee_fullname}
\bibliography{egbib}
}

\input{supplemental}

\end{document}

%% file: supplemental.tex
\appendix

\setcounter{page}{1}

\twocolumn[
\centering
\Large
\textbf{Efficient Classification of Very Large Images with Tiny Objects} \\
\vspace{0.5em}Supplementary Material \\
\vspace{1.0em}
] 
\appendix


\section{Memory Cost Analysis}
\label{sec: mem_cost}
In Memory requirements Section~\ref{sc:stage2}, we empirically analyze the memory requirements of the proposed {Zoom-In network} and the closely related {ATS} model.
Below we study the memory usage of these models using the colon cancer data.

\begin{figure*}[htb!]
\centering
\resizebox{1.0\linewidth}{!}{
\includegraphics[scale=0.55]{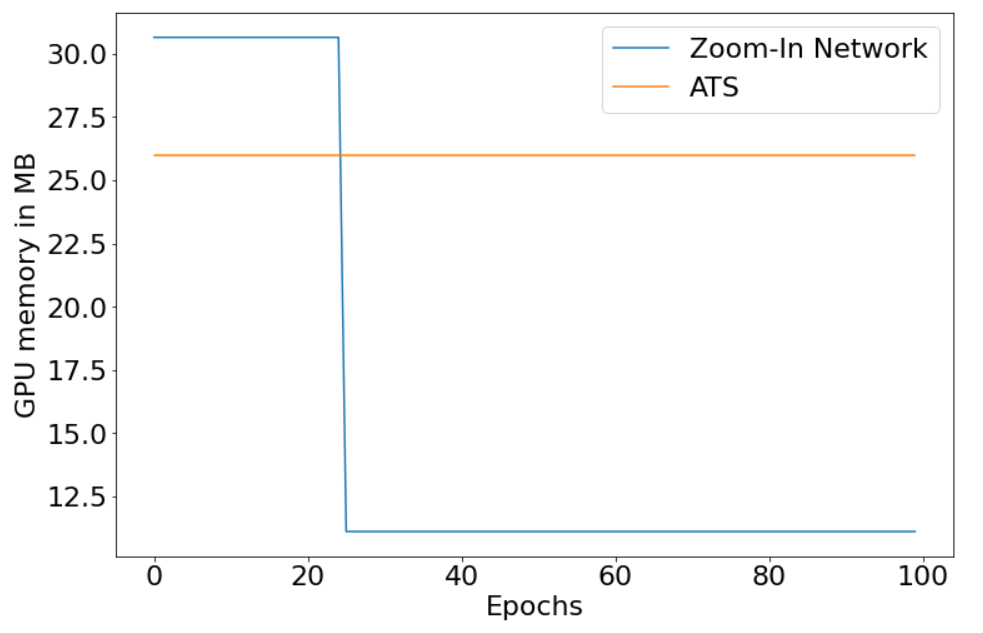}
\includegraphics[scale=0.55]{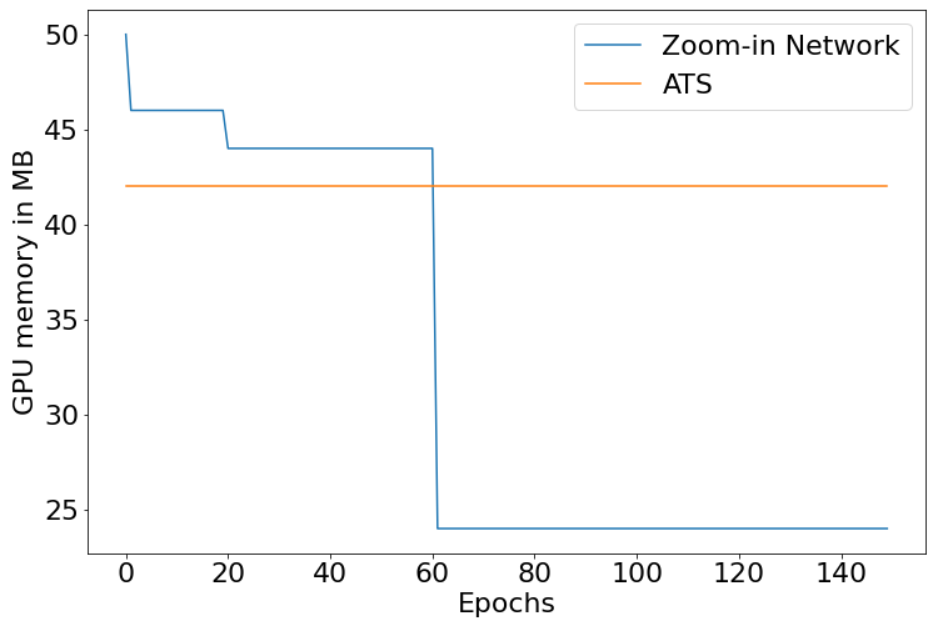}
\includegraphics[scale=0.55]{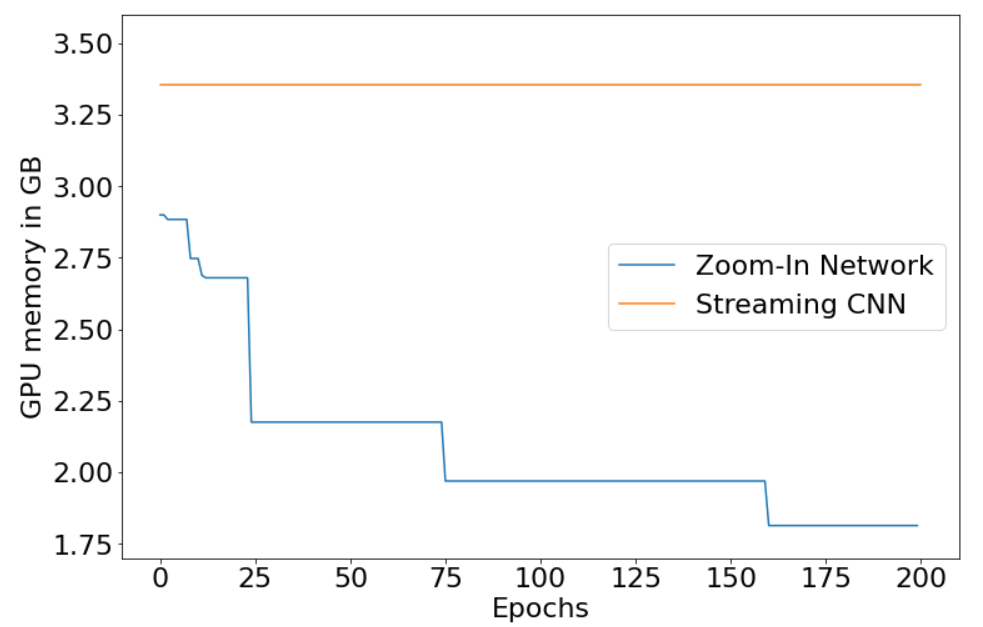}

}
\caption{
GPU memory usage (y-axis) versus training epoch (x-axis). We plot the GPU memory usage of the {Zoom-In network} and the one-stage {ATS} for the (left) colon cancer dataset, (middle) NeedleCamelyon dataset, and (right) Camelyon16 dataset.}
\label{fig:mem_all}
\end{figure*}

The memory usage for the one-stage {ATS} and the proposed two-stage model are $\mathcal{O}(s^{2}HW+Nh_2w_2)$ and $\mathcal{O}(s_{1}^{2}HW+N's_1^2s_2^2HW+Nh_2w_2)$ respectively.
Notations is inherited from the main paper, Zoom-In Network Section~\ref{sc:stage2}.
In the colon cancer experiment, $s=0.2$, $s_1 = 0.1$, $s_2=0.2$ and $N=10$.
Then, if we want the memory usage order for the two-stage model to be smaller than that of the one-stage model, we set $N'< 7.5$.
In the experiment, $N'<N/2$ when the two-stage model has converged.
It should be noted that in the beginning of training, the initial attention map $\alpha$ is approximately uniform because the weights of $a_\Theta(\cdot)$ are initialized at random.
The number selected tiles $N'$ is close to $N$, which implies that the memory consumption of the two-stage model is slightly larger than the one-stage model. 
However, after the attention network $a_\Theta(\cdot)$ is optimized, the number of selected tiles $N'$ drops dramatically.
At which time, the proposed two-stage model consumes much less memory than one-stage model as shown in Figure ~\ref{fig:mem_all}.
Note that for very large images (gigapixel in size), the number of selected tiles is much smaller than the size of sample space $|C|$, which means that even at the beginning of training, the two-stage model does not need to instantiate all tiles in an image and thus requires substantially less memory than the one-stage model (see right plot in Figure ~\ref{fig:mem_all}).

The above memory usage analysis of input entries can be reflected on the counts of FLOP, since the FLOPs is dominated by the size of the feature maps and model parameters, that is computed in the following way in agreement with \cite{Hard_attention, dong2017more}:
\begin{equation}
    FLOPs = C_{in}\times k^2 \times H_{out} \times W_{out} \times C_{out}
\end{equation}
where $C_{in}$ is the number of channels of the input tensor, $k^2$ is the size of the convolution kernels in this layer,  $H_{out}$, $W_{out}$ and $C_{out}$ are the heights, width and number of channels of the output tensor. 

\section{Derivations}
\label{sec: proofs}
Below we complete the proof for the equation~\eqref{eq:snorep} in the main paper, by showing the equivalence between $\mathbb{E}_{c'\sim b_{\Theta}(V(x, s_2, c))} [f_{\Theta}(T_{s_2}(T_{s_1}(x,c),c')) ]$ and $\sum_{c'\in C'} \sum_{i \neq c'} \beta_{c'}^c \frac{\beta_{i}^c}{1-\beta_{c'}^c}( \beta_{c'}^c f_\Theta(T_{s_2}(T_{s_1}(x,c),c') + (1-\beta_{c'}^c)f_\Theta(T_{s_2}(T_{s_1}(x,c),i))$ as follows,

\resizebox{.7\linewidth}{!}{
  \begin{minipage}{\linewidth}
\begin{align}\label{eq:snorep_proof}
    \begin{aligned}
   & \sum_{c'\in C'} \sum_{i \neq c'} \beta_{c'}^c \frac{\beta_{i}^c}{1-\beta_{c'}^c}( \beta_{c'}^c f_\Theta(T_{s_2}(T_{s_1}(x,c),c') +  (1-\beta_{c'}^c)f_\Theta(T_{s_2}(T_{s_1}(x,c),i) )  \\
  & = \sum_{c'\in C'}[\sum_{d \in C'} \beta_{c'}^c \frac{\beta_{d}^c}{1-\beta_{c'}^c}( \beta_{c'}^c f_\Theta(T_{s_2}(T_{s_1}(x,c),c') +  (1-\beta_{c'}^c)f_\Theta(T_{s_2}(T_{s_1}(x,c),d) ) -  \\
  &\beta_{c'}^c \frac{\beta_{c'}^c}{1-\beta_{c'}^c}( \beta_{c'}^c f_\Theta(T_{s_2}(T_{s_1}(x,c),c')  + (1-\beta_{c'}^c)f_\Theta(T_{s_2}(T_{s_1}(x,c),c'))] \\
  &  = \sum_{c'\in C'}[ \frac{{\beta_{c'}^c}^2}{1-\beta_{c'}^c} f_\Theta(T_{s_2}(T_{s_1}(x,c),c') +   \sum_{d \in C'}\beta_{d}^c\beta_{c'}^cf_\Theta(T_{s_2}(T_{s_1}(x,c),d) \\ & - \frac{{\beta_{c'}^c}^2}{1-\beta_{c'}^c} f_\Theta(T_{s_2}(T_{s_1}(x,c),c')]   \\
  &  = \sum_{c'\in C'}\beta_{c'}^c \sum_{d \in C'} \beta_{d}^cf_\Theta(T_{s_2}(T_{s_1}(x,c),d) 
  \\
  &= \sum_{d \in C'} \beta_{d}^cf_\Theta(T_{s_2}(T_{s_1}(x,c),d) \\
& = \mathbb{E}_{c'\sim b_{\Theta}(V(x, s_2, c))}  [f_{\Theta}(T_{s_2}(T_{s_1}(x,c),c')) ]. 
    \end{aligned}
\end{align}    
  \end{minipage}
}

\section{Supplementary Figures}
Here we present the supplementary figures mentioned in the main paper.

\begin{figure}[!htb]
\begin{center}\resizebox{1.0\linewidth}{!}{
\includegraphics[scale=0.6]{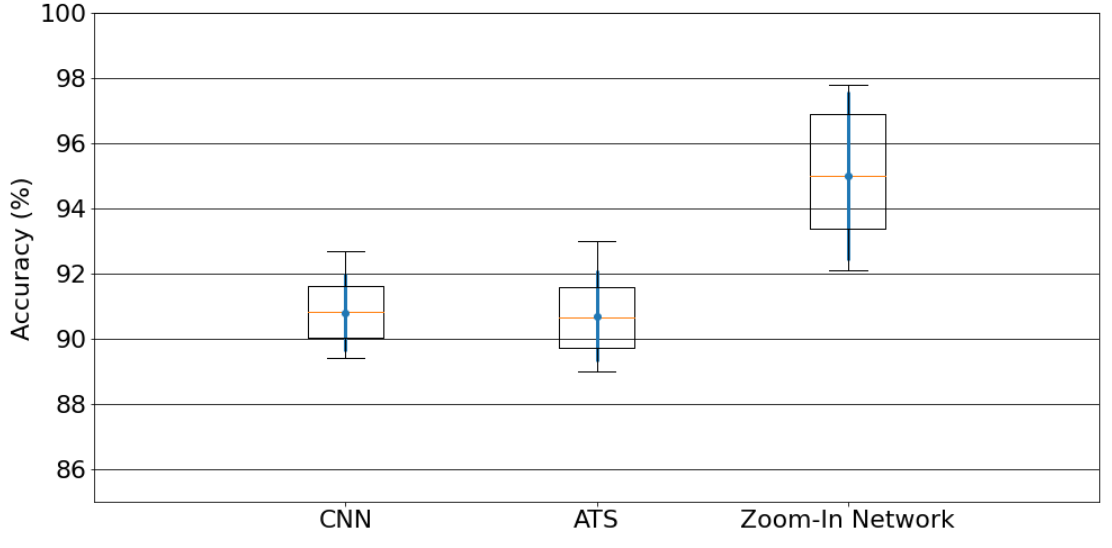}}\end{center}\vspace{-5mm}

   \caption{Standard deviations (error bars) for the results on the colon cancer dataset. Each error bar is obtained as the result of the 5 repetitions with the same training hyper-parameters and different random seeds. Standard deviations for CNN, ATS and {Zoom-In network} are $1.2\%$, $1.4\%$, $2.6\%$ respectively. We see that though the variation of the Zoom-In network is larger than for the CNN and ATS, the differences, accounting for variation are still significant in favor of the proposed model.}
\label{fig:error_bar}
\end{figure}

Figure~\ref{fig:error_bar} is the error bar plot for the results of colon cancer dataset.

\begin{figure}[!htb]
\begin{center}
\includegraphics[scale=0.38]{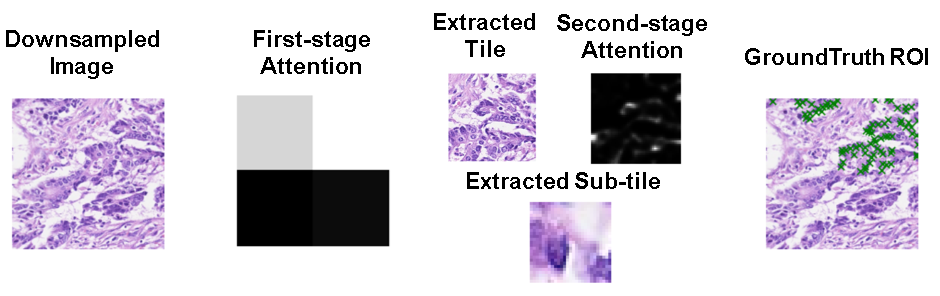}
\newline 
\\
\includegraphics[scale=0.38]{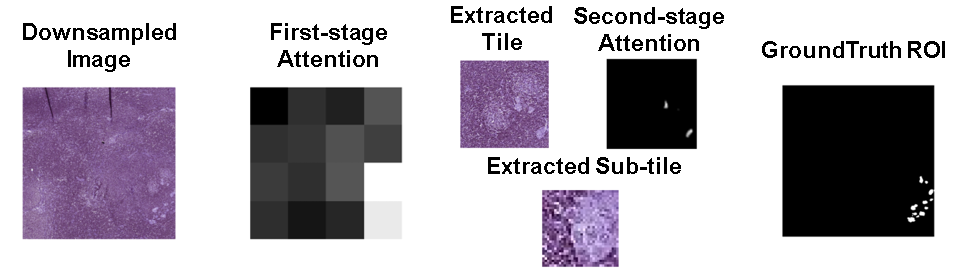}
\newline
\\
\includegraphics[scale=0.38]{figure/att_visual_traffic_sign.png}
\newline
\\
\includegraphics[scale=0.38]{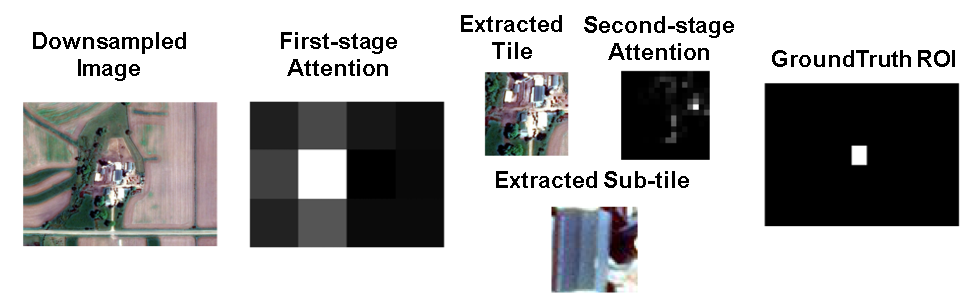}
\newline
\\
\includegraphics[scale=0.38]{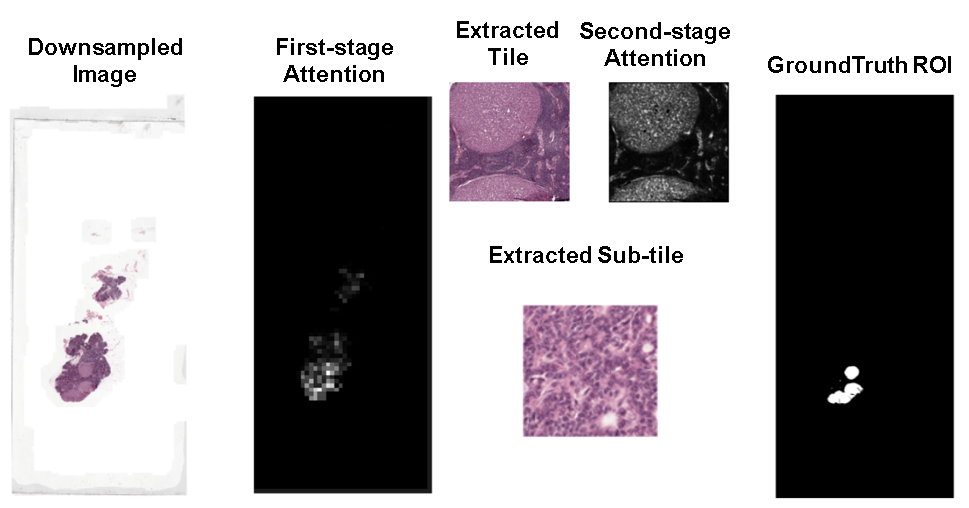}\end{center}
   \caption{Intermediate results of the proposed Zoom-In network. For each row, we exhibit the visualization of $a)$ colon cancer, $b)$ NeedleCamelyon, $c)$ Traffic Sign Recognition, $d)$ fMoW and $e)$ Camelyon16 datatsets. In each panel we show the downsampled original image, the ground truth ROI mask, the attention masks, the extracted tiles and sub-tiles with the highest first-stage and second-stage attention respectively. }
\label{fig:att_visual}
\end{figure}

Figure~\ref{fig:att_visual} illustrates the interpretability of the proposed attention model for the Colon Cancer, NeedleCamelyon, Traffic Sign Recognition, fMoW and Camelyon16 experiments.
We find that areas of high attention of the {Zoom-In network} and extracted ROI patches are highly consistent with the ground-truth, manually annotated, segmentation masks.

\begin{figure*}[!htb]
\begin{center}
\includegraphics[scale=0.52]{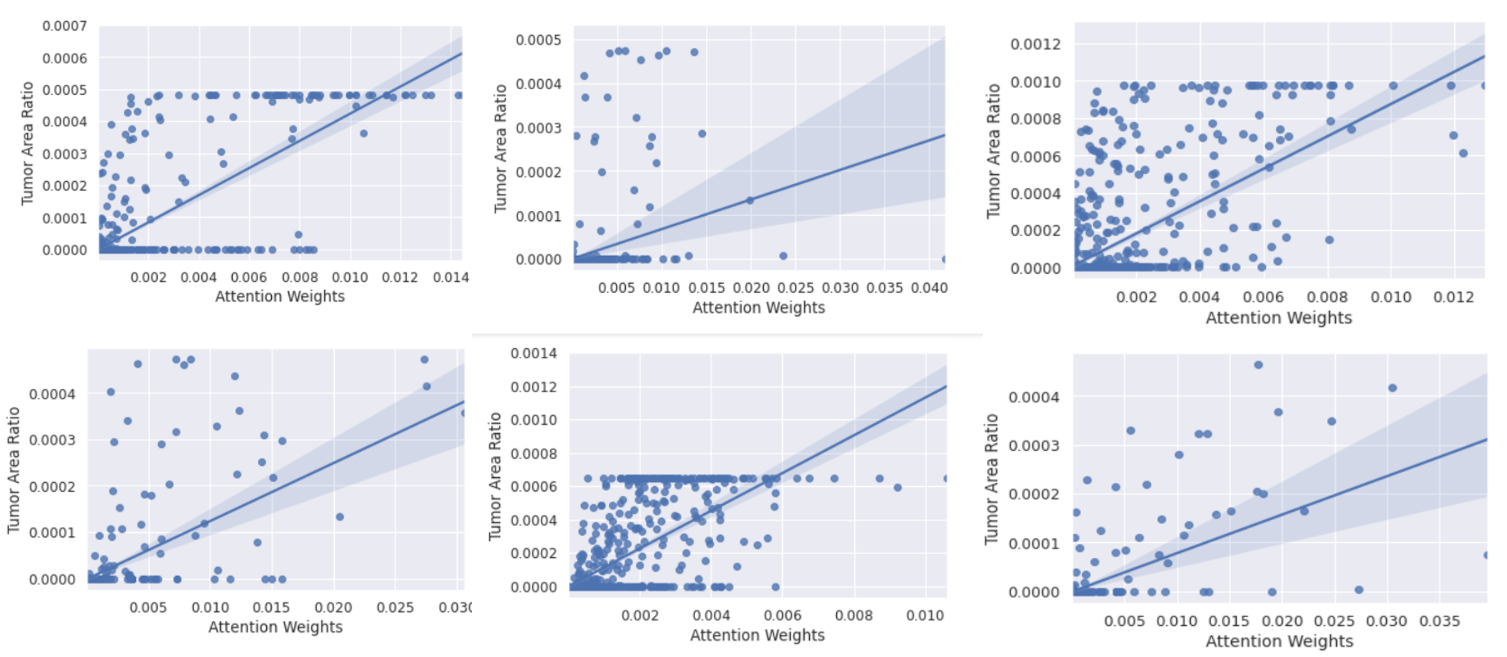}
\includegraphics[scale=0.5]{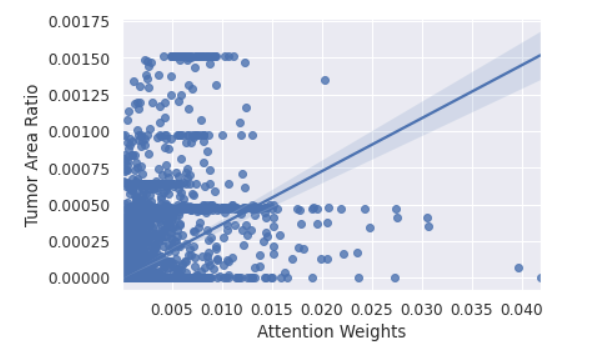}\end{center}
   \caption{Correlation of attention with ROI abundance, where the x-axis shows the attention weight $\alpha_c$ on each tile and the y-axis corresponds to the ratio of the ROI in a tile, {\em i.e.}, the franction of annotated ROI relative to the tile area.
   The plots are produced using images from the Camelyon16 dataset, for which detailed per-pixel annotations for cancer metastases are available. A linear regression fit is estimated for each image and overlayed on the plot to highlight the linear trend. (Top): 6 scatter plots created from 6 individual samples of the Camelyon16 test set. (Bottom): scatter plot created from the attention weight points and metastases ratio points in the whole dataset.}
\label{fig: CAM_corr}
\end{figure*}

Figure~\ref{fig: CAM_corr} quantitatively examines the quality of the attention of the {Zoom-In network}.
The plots illustrate the correlation of the attention weights generated by the Zoom-In network with the ground-truth metastases-to-tile ratio obtained from pixel annotations.
The Spearman correlation coefficient for all the tiles with pixel-level annotations is $\rho = 0.3570$, indicating a good agreement.
Interestingly, from the plot, we see that many tiles contain different size of ROIs but has the same magnitude of attention weights, indicating that the attention mechanism seems to attend to the presence of ROIs but not the proportion of ROIs in a tile.

\section{Details of Generating NeedleCamelyon Dataset}
\label{sec: generate needleCam}
We present the details of the NeedleCamelyon dataset not included in the main paper.
We first split the unique metastases in the original Camelyon16 dataset for positive samples in training set, validation set and test set.
The number of unique metastases region for training set, validation set and test set are $122$, $41$ and $41$, respectively.
Then we crop these metastases area to build the NeedleCamelyon dataset.
Negative images are taken by randomly cropping normal whole-slide images and filtering image crops that mostly contain background.
We ensure the class balance by sampling an equal amount of positive and negative crops.
Finally, we obtain $6000$ crops for training, $2000$ crops for validation and $2000$ crops for testing.
Each crop has a size of $1024 \times 1024$.

\section{Details of the Functional Map of the World Subset}
\label{sec: fmow subset}
In order to facilitate the experiments on fMoW dataset, we extract a subset from the original fMoW dataset. We randomly choose 10 categories from the original 63 categories. The chosen categories are airport, amusement, aquaculture, archaeological, barn, border, burial, car, construction and crop. Then, we randomly select 1,500 high-resolution images from the training set within the above classes as our training set and use all data belongs to the above categories in the validation set of orginal fMoW as our test set. 

\section{More Implementation Details of Attention $\alpha$ and $\beta$} 
\label{sec: attention implement}
Here, we describe more implementation details of attention $\alpha$ and $\beta$ not covered in Zoom-In Network Section~\ref{sec: methodology}.

In order to avoid additional computations on overlapped area in the image, each value in the first stage attention $\alpha$ represents the attention on a unique non-overlapped tile in the whole image. The tile sizes are $250 \times 250$ for colon cancer, $256 \times 256$ for NeedleCamelyon, $250 \times 250$ for Functional Map of the World and $3200 \times 3200$ for Camelyon16 dataset. For the second stage attention $\beta$, the attention maps are computed same as \cite{ATS}. Each attention value in $\beta$ is the attention of each input entry. 

\section{Complete Training Details}
\label{sec: complete training details}
Here, we describe the training details not included in Experiments Section\ref{sec: experiments}, such as learning rate, parameters of the optimizer, {\em etc}.

For the colon cancer dataset, we train our {Zoom-In network} using the Adam optimizer with a batch size of $5$, $\beta_1$ of $0.9$, $\beta_2$ of $0.999$ and a learning rate of $0.001$ for $100$ epochs.
We use $N$ of 10 and $\lambda$ of $1e^{-5}$.
The contrastive learning strategy kicks in after training for $10$ epochs. 

For NeedleCamelyon dataset, we set the learning rate to $0.0001$, $N$ to $30$ and training epochs to $150$.
Other settings are the same as for the colon cancer experiment.

For Traffic Sign Recognition, we set the learning rate to $0.001$, batch size to $32$, $N$ to $10$ and training epochs to $150$. Since the number of samples for each class is imbalanced, we use a weighted cross-entropy loss that the weights for class empty, 50 limit sign, 70 limit sign and 80 limit sign are $[0.1843, 2.3639, 1.5183, 3.1653]$. Other settings are conformed with the colon cancer experiment. 

For Functional Map of the World dataset, we set the learning rate to $0.0001$, batch size to $32$, $N$ to $30$ and training epochs to $200$. Other settings are the same as for the colon cancer experiment.

For Camelyon16 dataset, we set the learning rate to $0.0001$, $N$ to $100$ and training epochs to $200$.
Other settings are the same as for the colon cancer experiment.

All of our experiments ran on an NVIDIA TITAN Xp $12$GB with CUDA version $10.2$.
\section{Time/Memory-Accuracy Trade-off}
In our model, the main hyper-parameter that varies time and memory consumption is the sample size ($N$). Although the time/memory - accuracy trade-off can be inferred from the ablation study of sample size ($N$) in Table \ref{table: abs}, we show additional details concerning the time-memory trade-off in the following table:

\begin{table}[htb!]
\caption{Time/Memory-Accuracy Trade-off of Zoom-In Network compared with ATS in the colon cancer dataset experiment.} 
\centering 
\resizebox{1.0\columnwidth}{!}{
\begin{tabular}{c c c c c} 
  & Sample Sze & Accuracy & Memory & Time \\
Method & ($N$) &  (\%) &  (Mb) & (ms) \\
\hline 
	 & 5 & 93.2 &  2.43 & 3.04 \\ 
	Zoom-In Network & 10 & 78.0 & 2.55 & 3.20 \\
	 & 50 & 79.9 & 5.03 & 4.33  \\

	\hline
	ATS & 10 & 90.7 & 15.83	& 2.81\\ 

  & 50 & 90.7 & 25.77 & 4.03 \\  
\end{tabular}}
\label{table:time_mem_trade_off} 
\vspace{-5mm}
\end{table}
\section{Ablation Study}
\label{sec: as}
In Table~\ref{table: abs}, we present and ablation study to evaluate the effects of the entropy regularization $\lambda$, sample size $N$ and contrastive learning in our {Zoom-In network}.
We examine these hyper-parameters on the colon cancer dataset to show the effects of varying $N$ and $\lambda$, as well as to demonstrate the usefulness of the contrastive learning objective.
Then, we further justify the contribution of contrastive learning on NeedleCamelyon and Camelyon16 datasets a similar ablation strategy. 

\begin{table}[htb!]

\caption{Ablation study results of $\lambda$, $N$ and using contrastive learning. The first, seventh, and ninth rows are the standard hyper-parameter settings used in our experiments and the others are selected to show performance variations for different settings.} 
\centering 
\resizebox{1.0\columnwidth}{!}{
\begin{tabular}{c|cccc} 

    \multirow{2}{*}{Dataset}  &  Entropy  & Sample Size  & Contrastive  & Test Accuracy\\ 
    & Regularization ($\lambda$) & ($N$)& Learning & (\%)\\
    \hline
	\multirow{6}{*}{Colon Cancer} 
	& 1e-5 & 10 & Yes &  95.0 \\  
	& 0 & 10 & Yes &  94.0 \\  
	& 1.0 & 10 & Yes & 95.0\\
	& 1e-5 & 10 & No &  94.0 \\  
	& 1e-5 & 5 & Yes & 93.2 \\
	& 1e-5 & 50 & Yes & 96.0 \\ \hline
	\multirow{2}{*}{NeedleCamelyon} 
	& 1e-5 & 30 & Yes &  76.0 \\
	& 1e-5 & 30 & No & 74.3 \\ \hline 
	\multirow{2}{*}{Camelyon16}  
	 & 1e-5 & 100 & Yes & 81.3 \\
	 & 1e-5 & 100 & No & 80.6 \\
\end{tabular}}
\label{table: abs} 
\vspace{-4mm}
\end{table}

\section{Limitations: when there is no discriminative information at lower scales}
\label{sec: limit}
\begin{figure}[htb!]
\centering
\includegraphics[scale=0.2]{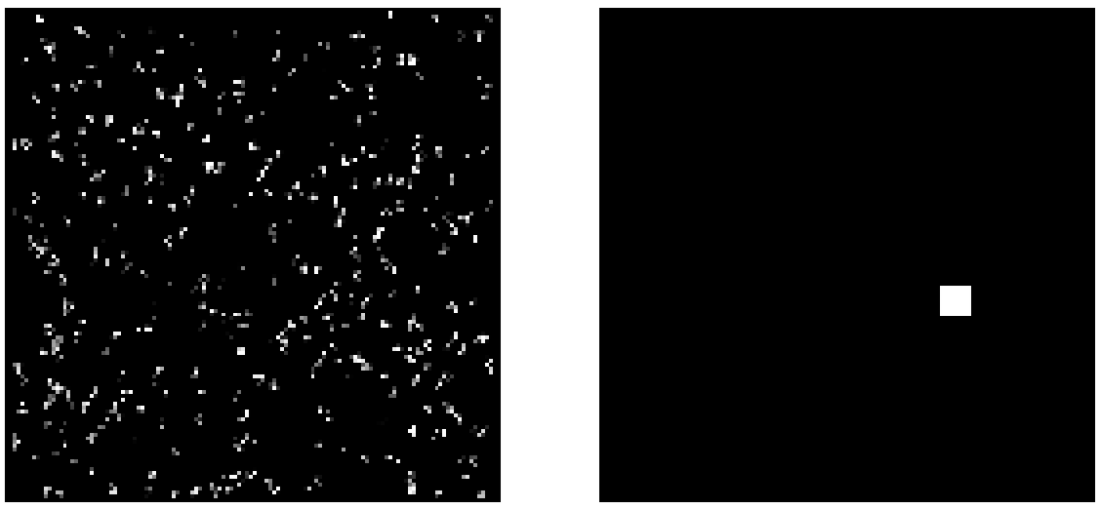}

\caption{Left: downsampled views of one image in NeedleMNIST dataset at scale $s_1=0.25$. Right: the corresponding ROI mask.}

\label{fig: NMNIST}
\end{figure}

We also examine the behavior of the proposed {Zoom-In network} on NeedleMNIST dataset introduced by \cite{pawlowski2019needles}.
In this dataset, images with a size of $1024 \times 1024$ are generated via randomly placing $401$ MNIST digits on a black image canvas.
The task for this dataset is to classify whether there is a digit $3$ in a given image.
The positive samples in this dataset contain only one digit $3$ and $400$ distracting digits, that is, any MNIST digits belong to a set of labels $\{0,1,2,4,5,6,7,8,9\}$.
The negative samples in this dataset have $401$ distracting digits.
We use the same split for training, validation and test set as described in \cite{pawlowski2019needles}.
The {Zoom-In network} fails to handle this dataset appropriately because the discriminative information is washed out when downsampling the view of the image. As shown Figure~\ref{fig: NMNIST}, we can hardly recognize the ROI of the downsampled view of NeedleMNIST images at scale $s_1 = 0.25$, which means that the {attention network} is unable to learn good attention weights that correlate with the ROI.
The test accuracy of the {Zoom-In network} is $0.503$, that is just slightly better than random guessing. There are a few possible solutions to address this problem: $i$) increasing the value of $N$, $s_1$ or $s_2$, at the cost of increased memory consumption; $ii$) pre-processing the images by a network pre-trained on ROI objects with labels (inferred from pixel-level annotations), in order to obtain feature maps with higher ROI-to-image ratio; and $iii$) randomly cropping the whole image so that some input images have higher ROI-to-image ratio.

\section{Model Components}

\begin{table}[htb!]

\caption{The architecture of the Zoom-In Network using a LeNet Structure.} 
\centering 
\begin{tabular}{|c| c|} 
 \multicolumn{2}{c}{$a_\Theta(\cdot)$}  \\
\hline

Layer & Type \\
\hline
1 & Conv(3, 1, 1, 8) + Tanh()  \\
2 & Conv(3, 1, 1, 8)  + Tanh()  \\
3 & Conv(3, 1, 1, 1)  + Tanh() \\
4 & GlobalAveragePooling2D() \\
5 & SoftMax() \\
\hline

\end{tabular}
\\
\begin{tabular}{|c| c|} 
\multicolumn{2}{c}{ }  \\
 \multicolumn{2}{c}{$b_\Theta(\cdot)$}  \\
\hline
Layer & Type \\
\hline
1 & Conv(3, 1, 1, 8) + Tanh()  \\
2 & Conv(3, 1, 1, 8)  + Tanh()  \\
3 & Conv(3, 1, 1, 1) + SoftMax()  \\
\hline

\end{tabular}
\\

\begin{tabular}{|c| c|} 
\multicolumn{2}{c}{ }  \\
 \multicolumn{2}{c}{$f_\Theta(\cdot)$}  \\
\hline
Layer & Type \\
\hline
1 & Conv(7, 1, 3, 32) + ReLU()  \\
2 & Conv(3, 1, 1, 32) + ReLU()  \\
3 & Conv(3, 1, 1, 32) + ReLU()  \\
4 & Conv(3, 1, 1, 32) + ReLU()  \\
5 & GlobalAveragePooling2D()  \\
\hline

\end{tabular}
\\
\begin{tabular}{|c| c|} 
\multicolumn{2}{c}{ }  \\
 \multicolumn{2}{c}{ $g_\Theta(\cdot)$}  \\
\hline
Layer & Type \\
\hline
1 & fc-$n_{class}$ \\
\hline
\end{tabular}
\\
\label{table: Le} 
\vspace{-4mm}
\end{table}

\begin{table}[htb!]

\caption{The architecture of the Zoom-In Network using a ResNet16 Structure.} 
\centering 
\begin{tabular}{|c| c|} 
 \multicolumn{2}{c}{$a_\Theta(\cdot)$}  \\
\hline

Layer & Type \\
\hline
1 & Conv(3, 1, 1, 8) + ReLU()  \\
2 & Conv(3, 1, 1, 16) + ReLU()  \\
3 & Conv(3, 1, 1, 32) + ReLU() \\
4 & Conv(3, 1, 1, 1) + ReLU() \\
5 & GlobalAveragePooling2D() \\
6 & SoftMax() \\
\hline

\end{tabular}
\\
\begin{tabular}{|c| c|} 
\multicolumn{2}{c}{ }  \\
 \multicolumn{2}{c}{$b_\Theta(\cdot)$}  \\
\hline
Layer & Type \\
\hline
1 & Conv(3, 1, 1, 8) + ReLU()  \\
2 & Conv(3, 1, 1, 16) + ReLU()  \\
3 & Conv(3, 1, 1, 32) + ReLU() \\
4 & Conv(3, 1, 1, 1) + SoftMax() \\
5 & SoftMax() \\
\hline

\end{tabular}
\\
\begin{tabular}{|c| c|} 
\multicolumn{2}{c}{ }  \\
 \multicolumn{2}{c}{$f_\Theta(\cdot)$}  \\
\hline
Layer & Type \\
\hline
1 & Conv(3, 1, 1)-32 + ReLU()  \\
2 & ResBlock(3, 1, 32)  \\
3 & ResBlock(3, 2, 32)  \\
4 & ResBlock(3,2, 32)  \\
5 & ResBlock(3,2, 32)  \\
6 & BatchNorm()+ReLU() \\
7 & GlobalAveragePooling2D()  \\
\hline

\end{tabular}

\begin{tabular}{|c| c|} 
\multicolumn{2}{c}{ }  \\
 \multicolumn{2}{c}{ $g_\Theta(\cdot)$}  \\
\hline
Layer & Type \\
\hline
1 & fc-$n_{class}$ \\
\hline
\end{tabular}
\label{table: ResNet16} 
\vspace{-4mm}
\end{table}

The components of the proposed Zoom-In network are summarized in Figure~\ref{Model_summary}. We consider the LeNet and ResNet16 architecutres for the feature function $f_\Theta(\cdot)$. The choice of LeNet is consistent to the one used in \cite{MIL, colon_cancer}.
The mapping $f_\Theta(\cdot)$ is implemented by a LeNet5-like architecture; consistent to the one used in \cite{MIL,colon_cancer}.
We also considered a ResNet architecture to show our zoom-in strategies is also compatible with modern network architectures. The details of each subnetwork is listed in Table ~\ref{table: Le} and Table ~\ref{table: ResNet16}. In the table, the convolutional layer is denoted as "Conv" and the kernel size, stride, padding and number of filters are provided in the following brackets. "fc" means the fully-connected layer and the output hidden units is provided after the dash. $n_{class}$ is the number of classes in the task that the model is solving. "Tanh", "ReLU" and "SoftMax" represent the non-linear functions. "GlobalAveragePooling2D"  is the global average pooling operation in the spatial dimension of the tensors, functioning the same as \url{https://www.tensorflow.org/api_docs/python/tf/keras/layers/GlobalAveragePooling2D}.  "ResBlock" is the standard ResNet block \cite{he2016deep}. In the brackets, we provide the kernel size, stride, and number of filters. 

The attention function $a_\theta(\cdot)$ in \eqref{eq:att} is a smaller neural network consisting of a three-layer convolutional network with 8 kernels and ReLu activations, followed by average pooling and a softmax activation to obtain the matrix of attention weights.
The attention function $b_\theta(\cdot)$ in \eqref{eq:att2} is defined similarly.

Finally, the classifier $g_\Theta(\cdot)$ is specified as a single fully connected layer with sigmoid activation.
The complete objective for the Zoom-In network is
\begin{align}
& \mathcal{L}(y,x;\Theta) = \label{eq:obj} \\
& \hspace{4mm} \mathcal{L}_{\rm ce}(y,x)+\mathcal{L}_{\rm con}(x,y=1)+\mathcal{L}_{\rm er}(\alpha) + \textstyle{\sum}_{c \in Q}\mathcal{L}_{\rm er} (\beta^c) , \notag
\end{align}
where $\mathcal{L}_{\rm ce}(y,x)$ is the cross-entropy loss for the image-level binary classification, $\mathcal{L}_{\rm con}(x,y=1)$ is the contrastive loss introduced above, and $\mathcal{L}_{\rm er}(\cdot)$ is the entropy regularization loss for attention matrices $\alpha$ and $\{\beta^c\}_c$.
The regularization term \cite{ATS,mnih2016asynchronous}, $\mathcal{L}_{\rm er}(p) = -\lambda H(p) = \lambda\sum_{i} p_i\log(p_i)$, where $H(\cdot)$ is the entropy of a discrete distribution and $\lambda$ is the trade-off coefficient, is included in the overall objective to prevent overly sparse attention matrices that may result from overfitting or early converging during training.

\section{Leveraging Pixel-Level Annotations}
\label{sec: pixel-level annotations}
In some tiny object image classification datasets, pixel-level annotations of the ROIs are provided; usually in the form of manually-created segmentation masks. In Table ~\ref{table:CAM16_result}, we also report the results of our model using the pixel-level annotations provided in the Camelyon16 dataset.
Below we describe how to leverage pixel-level annotations, when available, to further improve the performance of the proposed model.

\paragraph{Incorporating pixel-level annotations via pre-training}
A good attention function and a feature extractor can be obtained by training the model components as tile-level classifiers \cite{wang2016deep,qiu2018global}.
Specifically, assuming pixel-level annotations are available for (a subset of) the images, we can obtain patches consistent in size and scale with view $V(x,s_1)$ for $a_\Theta(V(x,s_1),)$ (of size $h\times w$), $V(x,s_2,c)$ for $b_\Theta(V(x,s_1,c))$ (of size $u\times v$), and with sub-tiles $T_{s_2}(T_{s_1}(x,c)c')$ for $f_\Theta(\cdot)$ (of size $h_2\times w_2$). 
Then, we can pair them with labels obtained from the pixel-level annotation so that the extracted patch is assigned $y=1$ is any of the annotation pixels is of a class different than background and $y=0$ if all the patch consists of background pixels.
Subsequently, we can proceed to pre-train $a_\Theta(V(x,s_1),)$,  $b_\Theta(V(x,s_1,c))$ and $f_\Theta(\cdot)$.
For the attention functions we convert their output to a scalar prediction using a global average pooling layer, and for $f_\Theta(\cdot)$ we use a single fully connected layer similar to $g_\Theta(\cdot)$, but whose parameters we discard after pre-training.

\paragraph{Incorporating pixel-level supervision}
Recently, \cite{song2018mask} introduced the body mask approach to guide the attention map, by adding a mean squared error (MSE) loss between the attention map for the positive class and the corresponding body segmentation mask to improve the model performance on person re-identification tasks, which is a detection task.
The segmentation mask (pixel-level annotation) is represented by a binary matrix of the same size as the original image.
If a pixel of the image is in a ROI (not background), the corresponding value in the mask is set to $1$, alternatively the value is set to $0$ (background).
Here, we optimize both attention networks using pixel-level annotations by adding MSE losses to the outputs of $a_\theta(V(x,s_1))$ and $b_\Theta(V(x,s_2,c))$.
The MSE losses used in our experiment are
\begin{align}
\begin{aligned}
\mathcal{L}_{\alpha}(\alpha) & = \textstyle{\sum}_{c \in C}||M(V(x, s_1), c)-\alpha_{c}||^2_2 , \\ 
\mathcal{L}_{\beta}(\beta^c) & = \textstyle{\sum}_{c' \in C'}||M(V(x, s_2), c')-\beta^c_{c'}||^2_2 ,
\end{aligned}
\end{align}
where $M(\cdot)$ is a function that returns the binary segmentation mask value for a specific view and location of the image, $\mathcal{L}_{\alpha}$ is the MSE loss for $\alpha=a_\theta(V(x,s_1))$ and $\mathcal{L}_{\beta}$ is the MSE loss for $\beta^c=b_\Theta(V(x,s_2,c))$.
These two losses are added to \eqref{eq:obj} when pixel-level annotations are available.




\section{Code and Data Availability}
\label{sec: code_data}
The source code of our project will be uploaded at \url{https://github.com/author_name/Zoom_in_network}. 

Colon cancer dataset can be downloaded at \url{https://github.com/MuniNihitha/cancer-detection/tree/master/data/CRCHistoPhenotypes_2016_04_28}.

The source code to reproduce NeedleCamelyon and NeedleMNIST dataset is at \url{https://github.com/facebookresearch/Needles-in-Haystacks}.

Traffic Sign dataset is avaliable at \url{https://www.cvl.isy.liu.se/research/trafficSigns/}.

Functional Map of the world(fMoW) dataset can be found at \url{https://github.com/fMoW/dataset}.

Camelyon16 dataset can be found at \url{https://camelyon16.grand-challenge.org/}.
